\documentclass[10pt,twocolumn,letterpaper]{article}

\usepackage[pagenumbers]{cvpr} 

\usepackage{graphicx}
\usepackage{amsmath}
\usepackage{amssymb}
\usepackage{booktabs}

\usepackage{enumitem}
\setlist{leftmargin=4.5mm}

\usepackage[breaklinks,colorlinks]{hyperref}

\usepackage[capitalize]{cleveref}
\crefname{section}{Sec.}{Secs.}
\Crefname{section}{Section}{Sections}
\Crefname{table}{Table}{Tables}
\crefname{table}{Tab.}{Tabs.}

\def\confName{CVPR}
\def\confYear{2022}

\begin{document}

\title{ABAW: Valence-Arousal Estimation, Expression Recognition, Action Unit Detection \& Multi-Task Learning Challenges}

\author{Dimitrios Kollias\\
Queen Mary University of London, UK\\
{\tt\small d.kollias@qmul.co.uk}
}
\maketitle

\begin{abstract}

This paper describes the third Affective Behavior Analysis in-the-wild (ABAW) Competition, held in conjunction with IEEE International Conference on Computer Vision and Pattern Recognition (CVPR), 2022. The 3rd ABAW Competition is a continuation of the Competitions held at ICCV 2021, IEEE FG 2020 and IEEE CVPR 2017 Conferences, and aims at automatically analyzing affect. This year the Competition encompasses four Challenges: i) uni-task Valence-Arousal Estimation, ii) uni-task Expression Classification, iii) uni-task Action Unit Detection, and iv) Multi-Task-Learning. All the Challenges are based on  a common benchmark database, Aff-Wild2, which is a large scale in-the-wild database and the first one to be annotated in terms of valence-arousal, expressions and action units.
In this paper, we present the four Challenges, with the utilized Competition corpora, we outline the evaluation metrics and present the baseline systems along with their obtained results.
\end{abstract}

\section{Introduction}

Affect recognition based on a subject’s facial expressions has been a topic of major research in the attempt to generate machines that can understand the way subjects feel, act and react. 
The problem of affect analysis and recognition constitutes a key issue in behavioural modelling, human computer/machine interaction and affective computing. 
There are a number of related applications spread across a variety of fields, such as medicine, health, or driver fatigue, monitoring, e-learning, marketing, entertainment, lie detection and law \cite{acharya2018automated,kim2016deep,nasser2019artificial,greene2016survey,zepf2020driver}.

In the past, due to the unavailability of large amounts of data captured in real-life situations, research has mainly focused on controlled environments. However, recently, social media and platforms have been widely used and large amount of data have become available. Moreover, deep learning has emerged as a means to solve visual analysis and recognition problems. Thus, major research has been given during the last few years to the development and use of deep learning techniques and deep neural networks \cite{lecun2015deep,goodfellow2016deep} in various applications, including affect recognition in-the-wild, i.e., in unconstrained environments. 
Moreover, apart from affect analysis and recognition, generation of facial affect is of great significance, in many real life applications, such as for synthesis of affect on avatars that interact with humans, in computer games, in augmented and virtual environments, in educational and learning contexts \cite{thies2016face2face,thies2018headon,pham2018generative,pumarola2018ganimation}. The ABAW Workshop exploits these advances and makes significant contributions for affect analysis, recognition and synthesis in-the-wild.

Ekman \cite{ekman2002facial} was the first to systematically study human facial expressions. His study categorizes the prototypical facial expressions, apart from neutral expression, into six classes representing anger, disgust, fear, happiness, sadness and surprise. Furthermore, facial expressions are related to specific movements of facial muscles, called Action Units (AUs). The Facial Action Coding System (FACS) was developed, in which facial changes are described in terms of AUs \cite{darwin1998expression}. 

Apart from the above categorical definition of facial expressions and related emotions, in the last few years there has been great interest in dimensional emotion representations, which are of great interest in human computer interaction and human behaviour analysis. Dimensional emotion representations are used to tag emotional states in continuous mode, usually in terms of the arousal and valence dimensions, i.e. in terms of how active or passive, positive or negative is the human behaviour under analysis \cite{frijda1986emotions,whissel1989dictionary,russell1978evidence}.

The third ABAW Competition, to be held in conjunction with the IEEE International Conference on Computer Vision and Pattern Recognition (CVPR), 2022 is a continuation of the first \cite{kollias2020analysing} and second \cite{kollias2021analysing} ABAW Competitions held in conjunction with  the IEEE  Conference  on  Face  and  Gesture Recognition (IEEE FG) 2021 and with the International Conference on Computer Vision (ICCV) 2022, respectively, which targeted dimensional (in terms of valence and arousal) \cite{deng2020multitask, li2021technical, zhang2020m, do2020affective,chen2017multimodal,weichi,deng2021towards,zhang2021prior,vu2021multitask,wang2021multi,zhang2021audio,xie2021technical,jin2021multi,antoniadis2021audiovisual,oh2021causal}, categorical (in terms of the basic expressions) \cite{kuhnke2020two,gera2020affect,dresvyanskiy2020audio,youoku2020multi,liu2020emotion,gera2021affect,mao2021spatial} and facial action unit analysis and recognition \cite{pahl2020multi,ji2020multi,kollias2021distribution,han2016incremental,kollias2019face,deng2020fau,saito2021action,vu2021multitask}. The third ABAW Competition contains four Challenges, which are based on the same in-the-wild database, (i) the uni-task Valence-Arousal Estimation Challenge; (ii) the uni-task Expression Classification Challenge (for the 6 basic expressions plus the neutral state plus the 'other' category that denotes expressions/affective states other than the 6 basic ones); (iii) the uni-task Action Unit Detection Challenge (for 12 action units); (iv) the Multi-Task Learning Challenge (for joint learning and predicting of valence-arousal, 8 expressions -6 basic plus neutral plus 'other'- and 12 action units). These Challenges produce a significant step forward when compared to previous events.
In particular, they use the Aff-Wild2 \cite{kollias2021analysing,kollias2020analysing,kolliasexpression,kollias2021affect,kollias2018aff2,kollias2018multi,kollias2021distribution,kollias2019face,kollias2018deep,zafeiriou,zafeiriou1}, the first comprehensive benchmark for all three affect recognition tasks in-the-wild: the Aff-Wild2 database extends the Aff-Wild \cite{kollias2018deep,zafeiriou,zafeiriou1}, with more videos and annotations for all behavior tasks. 

The remainder of this paper is organised as follows. We introduce the Competition corpora in Section \ref{corpora}, the Competition evaluation metrics in Section \ref{metrics}, the developed baseline, along with the obtained results in Section \ref{baseline}, before concluding in Section \ref{conclusion}.


\section{Competition Corpora}\label{corpora}


The third Affective Behavior Analysis in-the-wild (ABAW2) Competition relies on the Aff-Wild2 database, which is the first ever database annotated for all three main behavior tasks: valence-arousal estimation, action unit detection and expression classification. These three tasks constitute the basis of the four Challenges.

The Aff-Wild2 database, in all Challenges, is split into training, validation and test set. At first the training and validation sets, along with their corresponding annotations, are being made public to the participants, so that they can develop their own methodologies and test them. The training and validation data contain the videos and their corresponding annotation. Furthermore, to facilitate training, especially for people that do not have access to face detectors/tracking algorithms, we provide bounding boxes and landmarks for the face(s) in the videos (we also provide the aligned faces). At a later stage, the test set without annotations will be given to the participants. Again, we will provide bounding boxes and landmarks for the face(s) in the videos (we will also provide the aligned faces).

In the following, we provide a short overview of each Challenge's dataset and refer the reader to the original work for a more complete description. Finally, we describe the pre-processing steps that we carried out for cropping and aligning the images of Aff-Wild2. The cropped and aligned images have been utilized in our baseline experiments.

\subsection{Valence-Arousal Estimation Challenge}

This Challenge's corpora include $567$ videos in Aff-Wild2 that contain annotations in terms of valence and arousal. Sixteen of these videos display two subjects, both of which have been annotated. In total,  $2,816,832$ frames, with $455$ subjects, $277$ of which are male and $178$ female, have been annotated by four experts using the method proposed in \cite{cowie2000feeltrace}. Valence and arousal values range continuously in $[-1,1]$. 



\subsection{Expression Recognition Challenge}

This Challenge's corpora include $548$ videos in Aff-Wild2 that contain annotations in terms of the the 6 basic expressions, plus the neutral state, plus a category 'other' that denotes expressions/affective states other than the 6 basic ones. Seven of these videos display two subjects, both of which have been annotated. In total, $2,603,921$ frames, with $431$ subjects, $265$ of which are male and $166$ female, have been annotated by seven experts in a frame-by-frame basis. 



 

\subsection{Action Unit Detection Challenge}

This Challenge’s corpora include 547 videos that contain annotations in terms of 12 AUs, namely AU1, AU2, AU4, AU6, AU7, AU10, AU12, AU15, AU23, AU24, AU25 and AU26. Seven of these videos display two subjects, both of which have been annotated. In total, $2,603,921$ frames, with $431$ subjects, $265$ of which are male and $166$ female, have been annotated in a semi-automatic procedure (that involves manual and automatic annotations). The annotation has been performed in a frame-by-frame basis. Table \ref{au_distr} shows the name of the twelve action units that have been annotated and the action that they are associated with.

\begin{table}[h]
    \centering
        \caption{Distribution of AU annotations in Aff-Wild2}
    \label{au_distr}
\begin{tabular}{|c|c|}
\hline
  Action Unit \# & Action   \\   \hline 
    \hline    
   AU 1 & inner brow raiser  \\   \hline 
   AU 2 & outer brow raiser  \\   \hline  
   AU 4 & brow lowerer  \\   \hline 
   AU 6 & cheek raiser  \\   \hline  
   AU 7 & lid tightener \\   \hline 
   AU 10 & upper lip raiser \\   \hline 
   AU 12 & lip corner puller \\   \hline 
   AU 15 & lip corner depressor \\   \hline 
  AU 23 & lip tightener \\   \hline 
   AU 24 & lip pressor  \\   \hline 
   AU 25 & lips part  \\   \hline 
   AU 26 & jaw drop  \\   \hline 
\end{tabular}
\end{table}


 \subsection{Multi-Task-Learning Challenge}

For this Challenge’s corpora, we have created a static version of the Aff-Wild2 database, named s-Aff-Wild2. s-Aff-Wild2 contains selected-specific frames/images from Aff-Wild2. 
In total, 172,360 images are used that contain annotations in terms of valence-arousal; 6 basic expressions, plus the neutral state, plus the 'other' category; 12 action units (as described in the previous subsections).

\subsection{Aff-Wild2 Pre-Processing: Cropped \& Cropped-Aligned Images} \label{pre-process}


At first, all videos are splitted into independent frames. Then they are passed through the RetinaFace detector \cite{deng2020retinaface} so as to extract, for each frame, face bounding boxes and 5 facial landmarks. The images were cropped according the bounding box locations; then the images were provided to the participating teams.
The 5 facial landmarks (two eyes, nose and two mouth corners) were  used to perform similarity transformation. The resulting cropped and aligned images were additionally provided to the participating teams. Finally, the cropped and aligned images were utilized in our baseline experiments, described in Section \ref{baseline}.

All   cropped   and   cropped-aligned   images   were  resized   to $112 \times 112 \times 3$ pixel resolution and their intensity values were normalized  to  $[-1,1]$.

\section{Evaluation Metrics Per Challenge}\label{metrics}

Next, we present the metrics that will be used for assessing the performance of the developed methodologies of the participating teams in each Challenge.

\subsection{Valence-Arousal Estimation Challenge}

\noindent The performance measure is the average between the Concordance Correlation Coefficient (CCC) of valence and arousal. CCC evaluates the agreement between two time series (e.g., all video annotations and predictions) by scaling their correlation coefficient with their mean square difference. CCC takes values in the range $[-1,1]$; high values are desired. CCC is defined as follows:

\begin{equation} \label{ccc}
\rho_c = \frac{2 s_{xy}}{s_x^2 + s_y^2 + (\bar{x} - \bar{y})^2},
\end{equation}

\noindent
where $s_x$ and $s_y$ are the variances of all video valence/arousal annotations and predicted values, respectively, $\bar{x}$ and $\bar{y}$ are their corresponding mean values and $s_{xy}$ is the corresponding covariance value.

Therefore, the evaluation criterion for the  Valence-Arousal Estimation Challenge is:

\begin{equation} \label{va}
\mathcal{P}_{VA} = \frac{\rho_a + \rho_v}{2},
\end{equation}

\subsection{Expression Recognition Challenge}\label{evaluation}

\noindent The performance measure is the average F1 Score across all 8 categories (i.e., macro F1 Score). The $F_1$ score is a weighted average of the recall (i.e., the ability of the classifier to find all the positive samples) and precision (i.e., the ability of the classifier not to label as positive a sample that is negative). The $F_1$ score  takes values in the range $[0,1]$; high values are desired. The $F_1$ score is defined as:

\begin{equation} \label{f1}
F_1 = \frac{2 \times precision \times recall}{precision + recall}
\end{equation}

Therefore, the evaluation criterion for the  Expression Recognition Challenge is:

\begin{equation} \label{expr}
\mathcal{P}_{EXPR} = \frac{\sum_{expr} F_1^{expr}}{8}
\end{equation}

\subsection{Action Unit Detection Challenge}\label{evaluation2}

The performance measure is the average F1 Score across all 12 AUs (i.e., macro F1 Score). 
Therefore, the evaluation criterion for the  Action Unit Detection Challenge is:

\begin{equation} \label{au}
\mathcal{P}_{AU} = \frac{\sum_{au} F_1^{au}}{12}
\end{equation}

\subsection{Multi-Task-Learning Challenge}\label{mtl}

The performance measure is the sum of: the average CCC of valence and arousal; the average F1 Score of the 8 expression categories; the average F1 Score of the 12 action units (as defined above).
Therefore, the evaluation criterion for the  Multi-Task-Learning Challenge is:

\begin{align} \label{mtll}
\mathcal{P}_{MTL} &= \mathcal{P}_{VA} + \mathcal{P}_{EXPR} + \mathcal{P}_{AU} \nonumber \\
&=  \frac{\rho_a + \rho_v}{2} + \frac{\sum_{expr} F_1^{expr}}{8} + \frac{\sum_{au} F_1^{au}}{12}
\end{align}

\section{Baseline Networks and Results} \label{baseline}

All baseline systems rely exclusively on existing open-source machine learning toolkits to ensure the reproducibility of the results. All systems have been implemented in TensorFlow; training time was around six hours on a Titan X GPU, with a learning rate of $10^{-4}$ and with a batch size of 256. 

In this Section, we first describe the baseline systems developed for each Challenge and then we report their obtained results.

\begin{table*}[h]
\caption{Valence-Arousal Challenge Results on Aff-Wild2's validation set; evaluation criterion is the mean CCC of valence-arousal} 
\label{comparison_sota_va}
\centering
\begin{tabular}{ |c||c|c|c| }
 \hline
\multicolumn{1}{|c||}{\begin{tabular}{@{}c@{}} Baseline \end{tabular}}  &
\multicolumn{1}{c|}{\begin{tabular}{@{}c@{}} CCC-Valence  \end{tabular}} &
\multicolumn{1}{c|}{\begin{tabular}{@{}c@{}} CCC-Arousal\end{tabular}}
 &
\multicolumn{1}{c|}{\begin{tabular}{@{}c@{}} $\mathcal{P}_{VA}$  \end{tabular}}
\\ 
  \hline
 \hline
 
ResNet50  
&  \begin{tabular}{@{}c@{}} 0.31   \end{tabular} 
& \begin{tabular}{@{}c@{}}  0.17   \end{tabular} & 0.24 \\
\hline

\end{tabular}
\end{table*}

\begin{table*}[h]
\caption{Expression Challenge Results on Aff-Wild2's validation set;  evaluation criterion is the average F1 Score of 8 expressions} 
\label{comparison_sota_expr}
\centering
\scalebox{1.}{
\begin{tabular}{ |c||c| }
 \hline
\multicolumn{1}{|c||}{\begin{tabular}{@{}c@{}} Baseline \end{tabular}} & 
\multicolumn{1}{c|}{\begin{tabular}{@{}c@{}} $\mathcal{P}_{EXPR}$ \end{tabular}} 
\\ 
  \hline
 \hline
VGGFACE  & 0.23   \\
\hline

\end{tabular}
}

\end{table*}

\begin{table*}[h]
\caption{Action Unit Challenge Results on Aff-Wild2's validation set; evaluation criterion is the average F1 Score of 12 AUs} 
\label{comparison_sota_au}
\centering
\scalebox{1.}{
\begin{tabular}{ |c||c| }
 \hline
\multicolumn{1}{|c||}{\begin{tabular}{@{}c@{}} Baseline \end{tabular}} & 
\multicolumn{1}{c|}{\begin{tabular}{@{}c@{}} $\mathcal{P}_{AU}$ \end{tabular}} 
\\ 
  \hline
 \hline
VGGFACE  & 0.39   \\
\hline

\end{tabular}
}
\end{table*}

\begin{table*}[h]
\caption{Multi-Task-Learning Challenge Results on Aff-Wild2's validation set; evaluation criterion is the sum of each task's independent performance metric} 
\label{comparison_sota_mtl}
\centering
\scalebox{1.}{
\begin{tabular}{ |c||c| }
 \hline
\multicolumn{1}{|c||}{\begin{tabular}{@{}c@{}} Baseline \end{tabular}} & 
\multicolumn{1}{c|}{\begin{tabular}{@{}c@{}} $\mathcal{P}_{MTL}$ \end{tabular}} 
\\ 
  \hline
 \hline
VGGFACE  & 0.30   \\
\hline

\end{tabular}
}
\end{table*}

\subsection{Baseline Systems}

\paragraph{Valence-Arousal Estimation Challenge}

The baseline network is a ResNet one with 50 layers, pre-trained on ImageNet (ResNet50) and with a (linear) output layer that gives final estimates for valence and arousal.

\paragraph{Expression Recognition Challenge}

The baseline network is a VGG16 network with fixed (i.e., non-trainable) convolutional weights (only the 3 fully connected layers were trainable), pre-trained on the VGGFACE dataset and with an output layer equipped with softmax activation function which gives the 8 expression predictions.

\paragraph{Action Unit Detection Challenge}

The baseline network is a VGG16 network with  fixed convolutional weights (only the 3 fully connected layers were trained), pre-trained on the VGGFACE dataset and with an output layer equipped with sigmoid activation function which gives the 12 action unit predictions.

\paragraph{Multi-Task-Learning Challenge}

The baseline network is a VGG16 network with with fixed convolutional weights (only the 3 fully connected layers were trained), pre-trained on the VGGFACE dataset. The output layer consists of 22 units: 2 linear units that give the valence and arousal predictions; 8 units equipped with softmax activation function that give the expression predictions; 12 units equipped with sigmoid activation function that give the action unit predictions.

\subsection{Results}

Table \ref{comparison_sota_va} presents the CCC evaluation of valence and arousal predictions on the Aff-Wild2 validation set, of the baseline network (ResNet50). 
Table \ref{comparison_sota_expr} presents the performance, in the  Expression Classification Challenge, on the validation set of Aff-Wild2, of the baseline network (VGGFACE).
Table \ref{comparison_sota_au} presents the performance, in the  Action Unit Detection Challenge, on the validation set of Aff-Wild2, of the baseline network (VGGFACE).
Table \ref{comparison_sota_mtl} presents the performance, in the  Multi-Task-Learning Challenge, on the validation set of Aff-Wild2, of the baseline network (VGGFACE).

\section{Conclusion}\label{conclusion}
In this paper we have presented the third Affective Behavior Analysis in-the-wild Competition (ABAW) 2022 to be held in conjunction with IEEE CVPR 2022. This Competition  followed the first and second ABAW Competitions held in conjunction with  IEEE FG 2020 and ICCV 2021, respectively.  This Competition comprises  four Challenges targeting: i) uni-task valence-arousal estimation, ii) uni-task expression classification, iii) uni-task action unit detection and iv) multi-task-learning. The database utilized for this Competition has been derived from the Aff-Wild2, the first and large-scale database annotated for all these three behavior tasks.

{\small
\bibliographystyle{ieee_fullname}
\bibliography{egbib}
}

\end{document}